\documentclass{article} 
\usepackage{iclr2022_conference,times}

\usepackage{amsmath,amsfonts,bm}

\def\eqref#1{equation~\ref{#1}}

\def\1{\bm{1}}

\DeclareMathAlphabet{\mathsfit}{\encodingdefault}{\sfdefault}{m}{sl}
\SetMathAlphabet{\mathsfit}{bold}{\encodingdefault}{\sfdefault}{bx}{n}

\usepackage{hyperref}
\usepackage{url}
\usepackage[pdftex]{graphicx} 
\usepackage{booktabs} 
\usepackage{wrapfig}
\usepackage[font=small]{caption, subcaption}
\usepackage{color,soul} 

\title{A Revealing Large-Scale Evaluation of \\Unsupervised Anomaly Detection Algorithms}

\author{Maxime Alvarez\thanks{Equal contribution. Corresponding author: maxime.alvarez@usherbrooke.ca} , Jean-Charles Verdier\footnotemark[1] , D'Jeff K. Nkashama\footnotemark[1], \\
{\bf Marc Frappier, Pierre-Martin Tardif, Froduald Kabanza}\\
\ \\
GRIC, Université de Sherbrooke\\
Sherbrooke, QC, Canada\\
\texttt{\{maxime.alvarez,jean-charles.verdier,djeff.nkashama.kanda}\\
\texttt{\ marc.frappier,pierre-martin.tardif,froduald.kabanza\}@usherbrooke.ca}
}

\iclrfinalcopy 
\begin{document}

\maketitle

\begin{abstract}
Anomaly detection has many applications ranging from bank-fraud detection and cyber-threat detection to equipment maintenance and health monitoring. However, choosing a suitable algorithm for a given application remains a challenging design decision, often informed by the literature on anomaly detection algorithms. We extensively reviewed twelve of the most popular unsupervised anomaly detection methods. We observed that, so far, they have been compared using inconsistent protocols -- the choice of the class of interest or the positive class, the split of training and test data, and the choice of hyperparameters -- leading to ambiguous evaluations. This observation led us to define a coherent evaluation protocol which we then used to produce an updated and more precise picture of the relative performance of the twelve methods on five widely used tabular datasets. While our evaluation cannot pinpoint a method that outperforms all the others on all datasets, it identifies those that stand out and revise misconceived knowledge about their relative performances.
\end{abstract}

\section{Introduction}
A tenet of scientific publications is that the published results should be fair, unambiguous, and reproducible. However, there is an increasing awareness that this is not always the case in published machine learning research \cite{gorman2019we, agarwal2021deep, kadlec2017knowledge, fourure2021anomaly, musgrave2020metric, marie2021scientific, raff2019step}. To illustrate, sometimes methods are compared while using inconsistent hyparameter tunings \cite{kadlec2017knowledge, musgrave2020metric} or misleading metrics \cite{musgrave2020metric}, resulting in unfair comparative evaluations. Different choices for the anomaly class, differences in training protocol settings, and different proportions of anomalies in the test set were also identified as leading to inconsistent evaluations of machine learning models \cite{fourure2021anomaly}. Other studies, in deep reinforcement learning for example \cite{agarwal2021deep}, stress the importance of considering the statistical uncertainty uncured by comparing machine learning models on a small number of training runs, to ensure reliable performance evaluations.

This paper analyses twelve of the most popular unsupervised anomaly detection methods. We show that some of the issues raised by the abovementioned work also apply to the anomaly detection literature specifically. Anomaly detection algorithms attracted our attention because they are essential to a wide range of applications, ranging from bank-fraud detection \cite{zhu2021intelligent} and cyber-threat detection \cite{tan2011fast, schubert2014local} to equipment maintenance \cite{carvalho2019systematic} and health monitoring \cite{wei2018anomaly}.

An anomaly is an observation that deviates from what is deemed normal observations \cite{ruff2021unifying}. Anomaly detection algorithms are by nature classification algorithms and, like classification methods in general, are grouped into parametric and nonparametric approaches \cite{ruff2021unifying, kwon2019survey, chalapathy2019deep}. Parametric approaches assume a model defined by some parameters. Examples include decision-boundary learning methods, such as OC-SVM \cite{scholkopf1999support} and DeepSVDD \cite{ruff2018deep}, which learn a separation between normal and abnormal samples; reconstruction methods, such as MemAE \cite{gong2019memorizing} and DAE \cite{chen2018autoencoder}, which learn to reconstruct normal samples and classify samples that cannot be properly reconstructed as anomalous; and probabilistic methods, like DAGMM \cite{zong2018deep}, which learn the probability density function of the normal data. Nonparametric methods do not assume any parametric model. They include distance-based methods, like LOF \cite{breunig2000lof} and RecForest \cite{xu2021reconstruction}, which learn to identify anomalies by their distance to normal samples, given some distance-based metric.

The twelve anomaly detection methods we analyzed are: Deep Auto-Encoder \cite{chen2018autoencoder}, NeuTraLAD \cite{qiu2021neural}, DAGMM \cite{zong2018deep}, SOM-DAGMM \cite{chen2021multi}, DUAD \cite{li2021deep}, MemAE \cite{gong2019memorizing}, ALAD \cite{zenati2018adversarially}, DSEBM \cite{zhai2016deep}, DROCC \cite{goyal2020drocc}, DeepSVDD \cite{ruff2018deep}, LOF \cite{breunig2000lof} and OC-SVM \cite{scholkopf1999support}. After pointing to inconsistencies in published results for these methods, we propose a rigorous evaluation protocol and apply it to reevaluate those methods on five widely used tabular datasets: KDDCUP, NSL-KDD, CSE-CIC-IDS2018, Arrhythmia, and Thyroid. The new evaluation shows that some methods, previously shown to perform better than others, do not perform as well under our proposed protocol. We hope that the updated evaluation of anomaly detection algorithms will better inform future design choices of anomaly detection methods and future baselines for new method discoveries. We also believe that the adoption by authors of a coherent evaluation protocol like the one we propose will contribute to remedying the increasing concern of unfair, ambiguous, and difficult-to-reproduce experimental results in the machine learning literature.

\section{Issues with existing evaluations of machine-learning-based anomaly detection algorithms}
While reviewing the literature on machine-learning-based anomaly detection algorithms, we noticed inconsistencies in protocols used to evaluate and compare different algorithms, especially for the splitting between training and test datasets, the choice of performance metrics, and the threshold used to flag anomalies. We also observed ambiguity in the definition of the positive class (i.e., the class of interest) used for different models evaluations. As a result of these inconsistencies, it is difficult to make sense of the experimental evaluations from one paper to another. 

\textbf{Data splitting.} Different train-test data splits have been used while comparing the performances of different algorithms. One must usually decide on the training and test sets splitting. In addition, for anomaly detection, one must decide whether any of these two sets will contain normal data, abnormal data, or both. Whatever the decision, it should be made consistently when evaluating different algorithms. Otherwise, the comparisons of one paper to another are meaningless. It turns out that, in the literature, the data split decisions are indeed inconsistent. As an example, to compare different algorithms, some approaches split the data according to the following strategy \cite{zong2018deep, zenati2018adversarially, bergman2020classification}: the training dataset consists of 50\% of the normal data, whereas the test set consists of the remaining 50\% normal data plus anomalies. Based on this split, conclusions are made on the relative performance of different algorithms in the literature. We call this the ``Recycling strategy".

In contrast, \cite{zhai2016deep} start with a set containing both the normal data and anomalies, splits it evenly into two sets, then trains on a set that is one of the sets stripped of the anomalies, and tests it on the other set. We refer to this as the ``Discarding strategy". At first glance, the Recycling strategy and the Discarding strategy are similar in that training is done on normal data and testing is done on a mix of normal and abnormal data. The subtle difference here is that now the test set contains only half of the anomalies available in the original data. Considering that anomalies are rare -- they constitute a small percentage of the original dataset -- that subtle difference could have a significant impact on the testing performance \cite{fourure2021anomaly}. Put another way, the two strategies could lead to different conclusions when comparing anomaly detection algorithms on a dataset. For instance, \cite{zenati2018adversarially} refer to measurements made by previous authors using the Discarding strategy to compare against measurements made using the Recycling strategy in their paper. We later demonstrate that this is misleading, as one might expect. Regardless of the inconsistent comparison, since anomalous data is typically scarce, we argue that it should always be injected into the test set to evaluate the capacity of the model in detecting more anomalous signals. 

The two data strategies above say nothing about the proportion of anomalous and normal data in the test sets. Some authors fix that proportion to 50\%  --  making the test set balanced, thereby introducing a different test strategy \cite{goyal2020drocc}. We call it the ``Balanced test set strategy".  On the other hand, while the three strategies discussed so far train only on normal data, methods have also been proposed that train only on abnormal data. As an example, \cite{chen2021multi} use 50\% of the anomalies as the training set and the rest of the anomaly data plus normal data as the test set.

\textbf{Performance metrics and threshold}. Since most of the datasets in anomaly detection are imbalanced, precision, recall, and F1-score are commonly used metrics to measure model performance and benchmark models. These metrics are computed for a specific threshold in the anomaly detection task -- different thresholds may yield different metric values. The work in \cite{fourure2021anomaly} demonstrates that the F1-score with a fixed threshold can be artificially manipulated by increasing or decreasing the number of positive samples in the test set. Therefore, different anomaly ratios in the test set lead to biased comparisons. Moreover, the threshold can be another factor for manipulating performance measures. For instance, \cite{zong2018deep}, \cite{zenati2018adversarially} and \cite{bergman2020classification} set the threshold such that it returns the $\alpha^{th}$ percentile of anomaly scores, with $\alpha$ being the ratio of normal data in the test set; whereas others, like \cite{qiu2021neural}, search for an optimal threshold, which results in the best performance the model could achieve. Evidently, using different thresholds yields different results and performance comparison is only fair if the models are evaluated using their respective optimal thresholds.

\textbf{Class of interest (positive class)}. The choice of the positive class constitutes another source of ambiguity. For instance, \cite{zong2018deep} assign the positive class to the minority class (usually anomalous data) whereas, in \cite{goyal2020drocc} and \cite{chen2021multi}, the positive class denotes the majority class (normal data). However, with unbalanced data, the F1-score varies for class swapping \cite{chicco2020advantages}. Therefore, evaluating the performance of a model on the majority class gives it a clear advantage over those evaluated on the minority class.

\textbf{Implementation details}. Reproducibility in research allows double-checking findings and verifying whether they are reliable. It also facilitates the integration of recent findings when constructing new models. That said, reproducibility remains a challenge in the machine learning community, often due to important missing details in the description of models or the training procedure \cite{pineau2021improving}. In our literature review on anomaly detection algorithms, we noted similar issues. For instance, \cite{qiu2021neural} do not normalize values of features for some datasets, while \cite{zenati2018adversarially}, \cite{bergman2020classification}, \cite{zong2018deep} normalize values of attributes for all the datasets. The lack of implementation details may engender serious hurdles in the advancement of research in machine learning, in general; it reduces chances to reproduce results with sufficient certainty and impedes effective and consistent performance comparisons between different models.

\section{Proposed training and evaluation protocol}
In this section, we propose solutions for the issues identified in the previous section to ensure a fair, reliable, and consistent evaluation, and comparison of anomaly detection algorithms.

\subsection{Data split}
We propose to partition normal samples into a training and a test set following a 50-50 split using random subsampling. It could also be fair to use another split ratio, as long as all the anomalies are found exclusively within the test set. Given that anomalies are rare by nature and greatly influence the performance metrics, they should be included only in the test set. Also, training on normal data translates seamlessly to real-life applications where most of the data is assumed to be normal. However, for ablation studies, it can be informative to insert a small portion of the abnormal samples during training to study the algorithm's sensitivity to corruption, i.e. how the presence of anomalies in its training data affects its performance.

During our experiments, we trained and tested the models multiple times each. We kept the train-test constant across all runs however it has been pointed out by \cite{bouthillier2021accounting} that varying sources of randomness give a better estimation of the performances. Thus, it would be recommendable to shuffle all the data before splitting it into the training and test set at the beginning of each run.

\subsection{Class of interest}
In classification tasks, the class of interest, also called the positive class, is used as the basis for evaluation. Given the large class imbalance in anomaly detection, our protocol defines the minority class as the class of interest. By contrast, using the majority class gives overly optimistic scores and masks the poor performance on the minority class. In most cases, the minority class corresponds to the anomalies.

\subsection{Metrics}
We propose to use the following metrics to evaluate anomaly detection performance: F1-score, precision, recall, and area under the precision-recall curve (AUPR). AUPR was not used in any of the analyzed papers. As previously mentioned, research shows how F1-score and AUPR are sensitive to class imbalance \cite{jeni2013facing, tharwat2020classification} and can be manipulated by changing the anomaly ratio in the test set \cite{fourure2021anomaly}. Our protocol mitigates this issue by considering all the anomalies during testing. Thus, the anomaly ratio remains the same for all algorithms. While AUROC is unaffected by skewness in the class distribution, it provides an optimistic view by giving equal weights to predictions on both classes. The AUPR is more sensitive to predictions on the positive class (the anomalies), making it more informative for anomaly detection \cite{saito2015precision}. Also, AUPR and AUROC are not dependent on the choice of a specific threshold that can prevent comparability \cite{fourure2021anomaly}.

\subsection{Threshold}\
A threshold $\tau$ must be set to identify anomalies and compute the performance metrics such as F1-score, precision, and recall. Given a ratio of $\rho$ anomalous samples in the test set and the array S of generated scores on the entire test set, an intuitive strategy is to set $\tau$ at the $(1 - \rho)^{th}$ percentile of scores S. We expect the lowest (or highest, depending on the meaning of the score) of $(1 - \rho)^{th}$ percentile to contain the anomalies because they should generate the lowest (or highest scores). Another strategy is to find the optimal threshold, that is, the threshold that maximizes the F1-score. Such a threshold is typically located in the neighbourhood of the $(1 - \rho)^{th}$ percentile of scores S. We recommend the use of optimal thresholding for each model for a fair comparison.

As mentioned in the previous section, AUPR and AUROC are not dependent on the choice of a specific threshold. When comparing methods to apply to a problem where the anomaly ratio is unknown, these metrics are more informative than guessing the correct threshold. As such, these metrics are more attractive to industry practitioners.

\section{Experiments}
This section presents the datasets and the models used in our experiments along with implementation details. We then discuss the results obtained using the evaluation protocol suggested in the previous section.

\subsection{Datasets}
Commonly used datasets in unsupervised anomaly detection are considered for this task. Table \ref{dataset-info} summarizes the information of the different datasets.

\begin{table*}[!tbp]
    \vskip 0.15in
    \begin{center}
    \begin{small}
    \begin{sc}
    \begin{tabular}{l|c|c|c}
         Dataset &  Number of samples (N) & Number of features (D) & Anomaly ratio ($\rho$)\\
         \hline
         Arrhythmia & 452 & 274 & 0.1460\\
         CSE-CIC-IDS2018 & 16 232 944 & 83 & 0.1693\\
         KDD 10\% & 494 021 & 42 & 0.1969\\
         NSL-KDD & 148 517 & 42 & 0.4811\\
         Thyroid & 3772 & 6 & 0.0246
    \end{tabular}
    \end{sc}
    \end{small}
    \end{center}
    \caption{General information on the datasets.}
    \label{dataset-info}
    \vskip -0.1in
\end{table*}

\begin{itemize}
    \item \textbf{KDDCUP} is a network intrusion detection dataset widely used as a benchmark in the literature. We use the 10 percent version which uses only 10 percent of the original KDDCUP dataset. It contains 34 continuous and 7 categorical variables that are one-hot encoded. The four different attack scenarios (DOS, R2L, U2R, and probing) are combined into a single “attack” class. After manipulations, we drop the \texttt{num\_outbound\_cmds} and \texttt{is\_host\_login} columns because they both have a single value.
    
    \item \textbf{NSL-KDD}, provided by the Canadian Institute of Cybersecurity (CIC) \citep{nslkdd}, attempts to solve the inherent statistical flaws in KDDCUP (see \cite{tavallaee2009detailed} for more details) by removing most of the duplicate entries. The result is a much smaller training set with the same variables. The preprocessing steps is identical to KDDCUP. However, the column \texttt{is\_host\_login} is kept because it contains more than one value.
    
    \item \textbf{CIC-CSE-IDS2018}. This dataset is also provided by CIC. It simulates a complex enterprise network through virtual machines subject to seven different attack scenarios, namely Brute-force, Heartbleed, Botnet, DoS, DDoS, Web attacks, and infiltration of the network from inside \citep{cic-cse-ids2018}. Data cleaning for this dataset replicated the methodology described in \cite{leevy2021detecting}. Attacks are once again combined into a single class.
    
    \item \textbf{Thyroid}. This classification dataset taken from the ODDS repository \citep{thyroid} has three classes but, for the outlier detection task, only the hyperfunction class is treated as the outlier. The other two classes are treated as normal. The attributes are homogeneous and there is no missing or invalid data in this dataset.
    
    \item \textbf{Arrhythmia}. This multi-class classification dataset, also obtained from the ODDS repository \citep{arrhythmia}, combines multiple classes (3, 4, 5, 7, 8, 9, 14, and 15) to form the outlier class while the remaining classes are considered as the normal class. Like Thyroid, our data cleaning pipeline didn’t modify the original data.
\end{itemize}

Min-Max scaling was applied to all the features of the aforementioned datasets. Applying Min-Max scaling on one-hot encoded features has no effect.

\subsection{Models}
Our study compares the performance of 9 recent deep unsupervised learning algorithms tailored for anomaly detection, namely: DAGMM \cite{zong2018deep}, ALAD \cite{zenati2018adversarially}, MemAE \cite{gong2019memorizing}, DSEBM \cite{zhai2016deep} represented with its two alternative versions DBSEM-e and DBSEM-r, DROCC \cite{goyal2020drocc}, DeepSVDD \cite{ruff2018deep}, SOM-DAGMM \cite{chen2021multi}, DUAD \cite{li2021deep} and NeuTraLAD \cite{qiu2021neural}. They were chosen based on their performances and the diversity of approaches. We complement our comparison with two popular baseline methods: OC-SVM \cite{scholkopf1999support}, LOF \cite{breunig2000lof} and a vanilla Deep Auto Encoder (DAE) \cite{zhou2017anomaly}. 

Deep learning methods are implemented using PyTorch and optimized by the Adam algorithm with a learning rate of $1\mathrm{e}{-4}$, and training consists of 20 runs. Mini-batch sizes for Arrhythmia, Thyroid, KDD, NSL-KDD, and CSE-CIC-IDS2018 are set to 128, 128, 1024, 1024, and 1024 respectively. The author’s PyTorch versions of DeepSVDD, DROCC, and MemAE are integrated into our codebase. For ALAD, we had to convert the original TensorFlow codebase into PyTorch. Scikit-Learn’s LOF and OC-SVM implementations are used \cite{scikit-learn}. Finally, the remaining algorithms (DAGMM, DUAD, SOM-DAGMM, DSEBM, NeuTraLAD) are completely reimplemented because no public repository was made available by the authors. Our code is available on Github\footnote{\href{https://github.com/ireydiak/anomaly_detection_NRCAN}{ireydiak/anomaly\_detection\_NRCAN (github.com)}}.

\begin{table*}[!tbp]
\centering
\tiny

\begin{subtable}{\textwidth}
\centering
\begin{tabular}
{@{}llllllllllllll@{}}
\toprule & \multicolumn{3}{c}{KDDCUP 10} & \phantom{a} & \multicolumn{3}{c}{NSL-KDD} & \phantom{a} & \multicolumn{3}{c}{CSE-CIC-IDS2018} \\
\cmidrule{2-4} 
\cmidrule{6-8} 
\cmidrule{10-12} 
& Precision  & Recall & $F_1$ && Precision & Recall & $F_1$ && Precision & Recall & $F_1$  \\
\hline
ALAD        & 95.1$\pm$0.5 & 96.6$\pm$1.0 & 95.9$\pm$0.7 && 93.6$\pm$1.0 & 90.7$\pm$1.9 & 92.1$\pm$1.5 && 58.9$\pm$0.7 & 59.2$\pm$0.2 & 59.0$\pm$0.0\\
DAE         & 93.2$\pm$1.3 & 93.2$\pm$2.6 & 93.2$\pm$2.0 && \textbf{97.0$\pm$0.1} & 95.3$\pm$0.2 & \textbf{96.1$\pm$0.1} && 67.9$\pm$0.4 & 75.6$\pm$0.7 & 71.5$\pm$0.5 \\
DAGMM       & 93.6$\pm$0.9 & 98.4$\pm$1.9 & 95.9$\pm$1.4 && 89.3$\pm$5.5 & 81.8$\pm$9.0 & 85.3$\pm$7.4 && 48.4$\pm$4.1 & 65.9$\pm$7.3 & 55.8$\pm$5.3 \\
DeepSVDD    & 90.8$\pm$2.0  & 87.6$\pm$2.0  & 89.1$\pm$2.0  && 89.4$\pm$2.0  & 89.2$\pm$2.0  & 89.3$\pm$2.0 && 20.7$\pm$11 & 20.8$\pm$11 & 20.8$\pm$11 \\
DROCC       & 84.0$\pm$0.0 & 99.6$\pm$0.0 & 91.1$\pm$0.0 && 90.4$\pm$0.0 & 90.5$\pm$0.0 & 90.4$\pm$0.0 && 29.6$\pm$0.0 & \textbf{99.6$\pm$0.0} & 45.6$\pm$0.0 \\
DSEBM-e     & 95.7$\pm$0.1 & 97.6$\pm$0.1 & 96.6$\pm$0.1 && 95.5$\pm$0.1 & 93.7$\pm$0.1 & 94.6$\pm$0.1 && 45.1$\pm$0.7 & 42.7$\pm$0.8 & 43.9$\pm$0.8 \\
DSEBM-r     & \textbf{96.6$\pm$0.1} & 99.4$\pm$0.1 & \textbf{98.0$\pm$0.1} && 96.2$\pm$.0.1 & 94.9$\pm$0.1 & 95.5$\pm$0.1 && 42.2$\pm$0.1 & 39.3$\pm$0.1 & 40.7$\pm$0.1 \\
DUAD        & 94.0$\pm$0.7 & 99.1$\pm$1.4 & 96.5$\pm$1.0 && 96.0$\pm$0.1 & 93.2$\pm$0.3 & 94.5$\pm$0.2 && 68.1$\pm$3.5 & 75.8$\pm$2.4 & \textbf{71.8$\pm$2.7} \\
MemAE       & 93.0$\pm$1.2 & 97.1$\pm$2.2 & 95.0$\pm$1.7 && 96.0$\pm$0.0 & 95.1$\pm$0.1 & 95.6$\pm$0.0 && 60.8$\pm$0.1 & 59.0$\pm$0.2 & 59.9$\pm$0.1 \\
NeuTraLAD   & 93.1$\pm$0.3 & \textbf{99.7$\pm$0.1} & 96.4$\pm$0.2 && 96.5$\pm$0.4 & \textbf{95.6$\pm$0.2} & 96.0$\pm$0.1 && 54.6$\pm$8.2 & 65.6$\pm$9.8 & 59.5$\pm$8.9 \\
SOM-DAGMM   & 95.7$\pm$0.7 & 99.8$\pm$0.2 & 97.7$\pm$0.3 && 94.4$\pm$1.0 & 96.8$\pm$0.8 & 95.6$\pm$0.3 && 48.6$\pm$1.2 & 40.3$\pm$0.9 & 44.1$\pm$1.1 \\
LOF         & 93.0$\pm$0.0 & 97.2$\pm$0.0 & 95.1$\pm$0.0 && 88.6$\pm$0.0 & 93.6$\pm$0.0 & 91.1$\pm$0.0 & & 75.6$\pm$0.0 & 55.1$\pm$0.0 & 63.8$\pm$0.0 \\
OC-SVM      & 94.2$\pm$0.0 & 99.4$\pm$0.0 & 96.7$\pm$0.0 && 91.5$\pm$0.0 & 94.5$\pm$0.0 & 93.0$\pm$0.0 && \textbf{92.5$\pm$0.0} & 30.6$\pm$0.0 & 45.4$\pm$0.0 \\
\bottomrule
\end{tabular}
\caption{Performance metrics on cybersecurity datasets.}
\end{subtable}

\begin{subtable}{\textwidth}
\centering
\begin{tabular}
{@{}lllllllll@{}}
\\
\toprule & \multicolumn{3}{c}{Arrhythmia} & \phantom{abc} & \multicolumn{3}{c}{Thyroid}\\
\cmidrule{2-4} \cmidrule{6-8} & Precision & Recall & $F_1$ && Precision & Recall & $F_1$ \\
\hline
ALAD        & 59.5$\pm$0.1 & 55.5$\pm$0.8 & 57.4$\pm$0.4 &&  61.5$\pm$0.9 & 77.4$\pm$0.8 & 68.6$\pm$0.5\\ 
DAE         & 62.1$\pm$2.0 & 60.9$\pm$2.2 & 61.5$\pm$2.5 &&  54.4$\pm$4.4 & 65.5$\pm$5.7 & 59.0$\pm$1.5\\
DAGMM       & 51.4$\pm$3.9 & 50.0$\pm$6.1 & 50.6$\pm$4.7 &&   51.3$\pm$7.2 & 46.3$\pm$8.8 & 48.6$\pm$8.0\\
DeepSVDD    & 56.4$\pm$2.0 & 54.7$\pm$3.0 & 55.5$\pm$3.0 &&  13.0$\pm$13 & 24.7$\pm$30 & 13.1$\pm$13\\
DROCC       & 26.3$\pm$3.6 & 61.8$\pm$15 & 35.8$\pm$2.6 &&  51.8$\pm$9.0 & 77.6$\pm$11 & 62.1$\pm$10\\
DSEBM-e     & 61.9$\pm$1.0 & 58.1$\pm$1.6 & 59.9$\pm$1.0 &&  26.0$\pm$0.9 & 22.0$\pm$0.6 & 23.8$\pm$0.7\\
DSEBM-r     & 61.8$\pm$1.1 & 58.4$\pm$1.3 & 60.1$\pm$1.0 &&  25.8$\pm$0.3 & 21.8$\pm$0.4 & 23.6$\pm$0.4\\
DUAD        & 58.6$\pm$0.4 & 63.2$\pm$1.2 & 60.8$\pm$0.4 &&  12.3$\pm$2.0 & 19.1$\pm$1.6 & 14.9$\pm$5.5 \\
MemAE       & \textbf{63.1$\pm$2.1} & 62.1$\pm$1.5 & 62.6$\pm$1.6 &&  53.4$\pm$0.5 & 59.1$\pm$0.5 & 56.1$\pm$0.9\\
NeuTraLAD     & 58.5$\pm$5.5 & 63.6$\pm$5.3 & 60.7$\pm$3.7 &&  68.9$\pm$0.7 & \textbf{78.5$\pm$0.5} & \textbf{73.4$\pm$0.6} \\
SOM-DAGMM   & 51.0$\pm$5.9 & 53.2$\pm$7.5 & 51.9$\pm$5.9 && 61.2$\pm$11 & 49.0$\pm$14 & 52.7$\pm$12\\
LOF         & 57.1$\pm$0.0 & 66.7$\pm$0.0 & 61.5$\pm$0.0 &&  63.0$\pm$0.0 & 75.2$\pm$0.0 & 68.6$\pm$0.0\\
OC-SVM      & 57.3$\pm$0.0 & \textbf{71.2$\pm$0.0} & \textbf{63.5$\pm$0.0} &&  \textbf{69.6$\pm$0.0} & 66.6$\pm$0.0 & 68.1$\pm$0.0\\
\bottomrule
\end{tabular}
\caption{Performance metrics on medical datasets.}
\end{subtable}
\caption{Average Precision, Recall, and F1-Score (all with standard deviation) of the twelves models trained exclusively on the normal data.}
\label{f1_results_table}
\end{table*}

\begin{table*}[!tbp]
\tiny

\begin{subtable}{\textwidth}
\centering
\begin{tabular}
{@{}lcccccccccccc@{}}
\toprule & \multicolumn{2}{c}{KDD10} & 
\phantom{abc} & \multicolumn{2}{c}{NSL-KDD} &
\phantom{abc} & \multicolumn{2}{c}{CSE-CIC-IDS2018} \\
\cmidrule{2-3} \cmidrule{5-6} \cmidrule{8-9}  &
AUROC & AUPR &&
AUROC & AUPR &&
AUROC & AUPR \\
\hline
ALAD        & 99.0$\pm$0.2      & 95.3$\pm$1.1      && 93.9$\pm$1.8         & 94.8$\pm$1.7      && 85.6$\pm$2.1         & 61.5$\pm$5.8 \\
DAE         & 98.2$\pm$0.0      & 94.7$\pm$0.1      && 98.5$\pm$0.2         & \textbf{99.2$\pm$0.1}&& \textbf{88.2$\pm$1.8}      & 68.9$\pm$2.5\\
DeepSVDD    & \textbf{99.4$\pm$0.2} & 97.1$\pm$1.0  && 93.1$\pm$3.0         & 95.3$\pm$1.0       && 62.0$\pm$7.4        & 24.2$\pm$8.0 \\
DROCC       & 97.5$\pm$0.0      & 93.2$\pm$0.0      && 94.4$\pm$0.0         & 97.2$\pm$0.0      && 51.1$\pm$0.0         & 29.6$\pm$0.0 \\
DSEBM-e     & 98.6$\pm$0.1      & 93.9$\pm$0.5      && 98.0$\pm$0.0         & 99.0$\pm$0.0      && 70.7$\pm$0.2         & 38.1$\pm$0.1 \\
DSEBM-r     & 99.0$\pm$0.0      & 95.6$\pm$0.2      && 98.3$\pm$0.0         & 99.1$\pm$0.0      && 72.7$\pm$1.7         & 39.8$\pm$1.3 \\
DAGMM       & 99.1$\pm$0.3      & \textbf{97.3$\pm$0.6} && 94.0$\pm$3.6     & 96.5$\pm$2.0      && 66.7$\pm$8.3         & 50.8$\pm$9.5 \\
DUAD        & 98.3$\pm$1.0      & 93.2$\pm$3.5      && 97.4$\pm$0.1         & 98.6$\pm$0.1      && 82.9$\pm$2.3         & 53.0$\pm$0.1 \\
MemAE       & 98.2$\pm$0.2      & 94.7$\pm$0.6      && 97.9$\pm$0.1         & 98.9$\pm$0.1      && 65.8$\pm$2.0         & 56.6$\pm$2.0  \\
NeuTraLAD   & 98.8$\pm$0.1      & 97.0$\pm$0.1      && \textbf{98.7$\pm$0.1} & \textbf{99.2$\pm$0.1} && 81.5$\pm$5.6    & 59.6$\pm$6.9\\
SOM-DAGMM   & 98.9$\pm$0.2      & 95.8$\pm$1.3      && 98.6$\pm$0.2         & 99.2$\pm$0.1      && 67.9$\pm$6.4               & 53.2$\pm$2.4 \\
LOF         & 91.1$\pm$0.0      & 89.9$\pm$0.0      && 91.1$\pm$0.0         & 89.9$\pm$0.0      && 83.4$\pm$0.0         & \textbf{72.7$\pm$0.0} \\
OC-SVM      & 98.8$\pm$0.0      & 94.9$\pm$0.0      && 96.5$\pm$0.0         & 96.4$\pm$0.0      && 64.6$\pm$0.0         & 48.2$\pm$0.0\\
\bottomrule
\end{tabular}
\caption{AUROC and AUPR scores on cybersecurity datasets.}
\end{subtable}
\hfill
\begin{subtable}{\textwidth}
\centering
\begin{tabular}
{@{}lcccccccccccc@{}}
\\
\toprule & \multicolumn{2}{c}{Arrhythmia} & 
\phantom{abc} & \multicolumn{2}{c}{Thyroid} \\
\cmidrule{2-3} \cmidrule{5-6}  &
AUROC & AUPR &&
AUROC & AUPR \\
\hline
ALAD        & 78.7$\pm$1.3 & 62.0$\pm$1.6       && 85.7$\pm$3.9 & 59.9$\pm$4.8 \\
DAE         & \textbf{81.7$\pm$0.6} & \textbf{67.5$\pm$0.9}       && 95.1$\pm$0.8 & 54.2$\pm$3.1 \\
DAGMM       & 68.9$\pm$2.9 & 45.8$\pm$5.2       && 84.3$\pm$2.6 & 37.8$\pm$5.9 \\
SOM-DAGMM   & 70.3$\pm$5.0 & 48.7$\pm$6.9       && 85.0$\pm$7.1 & 46.7$\pm$13 \\
DUAD        & 81.2$\pm$0.4 & 66.8$\pm$0.4       && 43.0$\pm$0.3 & 4.6$\pm$0.5\\
MemAE       & 80.9$\pm$0.1 & \textbf{67.5$\pm$0.1}       && 89.8$\pm$4.6 & 32.7$\pm$7.0  \\
DeepSVDD    & 79.4$\pm$0.8 & 62.5$\pm$0.6       && 83.7$\pm$14 & 51.6$\pm$18\\
DROCC       & 64.0$\pm$4.3 & 30.8$\pm$3.9       && 95.6$\pm$3.8 & 68.9$\pm$15 \\
DSEBM-e     & 80.1$\pm$0.2 & 67.0$\pm$1.0        && 85.1$\pm$0.2 & 24.3$\pm$0.6\\
DSEBM-r     & 80.4$\pm$2.2 &  66.9$\pm$2.0      && 85.6$\pm$0.1 & 25.0$\pm$0.6 \\
NeuTraLAD   & 78.9$\pm$2.6 & 63.4$\pm$3.3       && \textbf{98.2$\pm$2.2} & \textbf{73.9$\pm$2.9} \\
LOF         & 81.3$\pm$0.0 & 67.0$\pm$0.0       && 97.2$\pm$0.0 & 72.2$\pm$0.0 \\
OC-SVM      & 80.0$\pm$0.0 & 64.0$\pm$0.0       && 96.9$\pm$0.0 & 61.4$\pm$0.0 \\
\bottomrule
\end{tabular}
\caption{AUROC and AUPR  on  medical datasets}
\end{subtable}

\caption{Average AUROC and AUPR (all with standard deviation) of the twelves models trained exclusively on the normal data.}
\label{auroc_results_table}
\end{table*}

\subsection{Results}
We now highlight the different results using our experimental protocol. Recall, precision, and F1-score are presented in Table \ref{f1_results_table}. Table \ref{auroc_results_table} displays the AUPR results.

\begin{figure}
     \centering
     \begin{subfigure}[b]{0.45\linewidth}
         \centering
         \includegraphics[width=\linewidth]{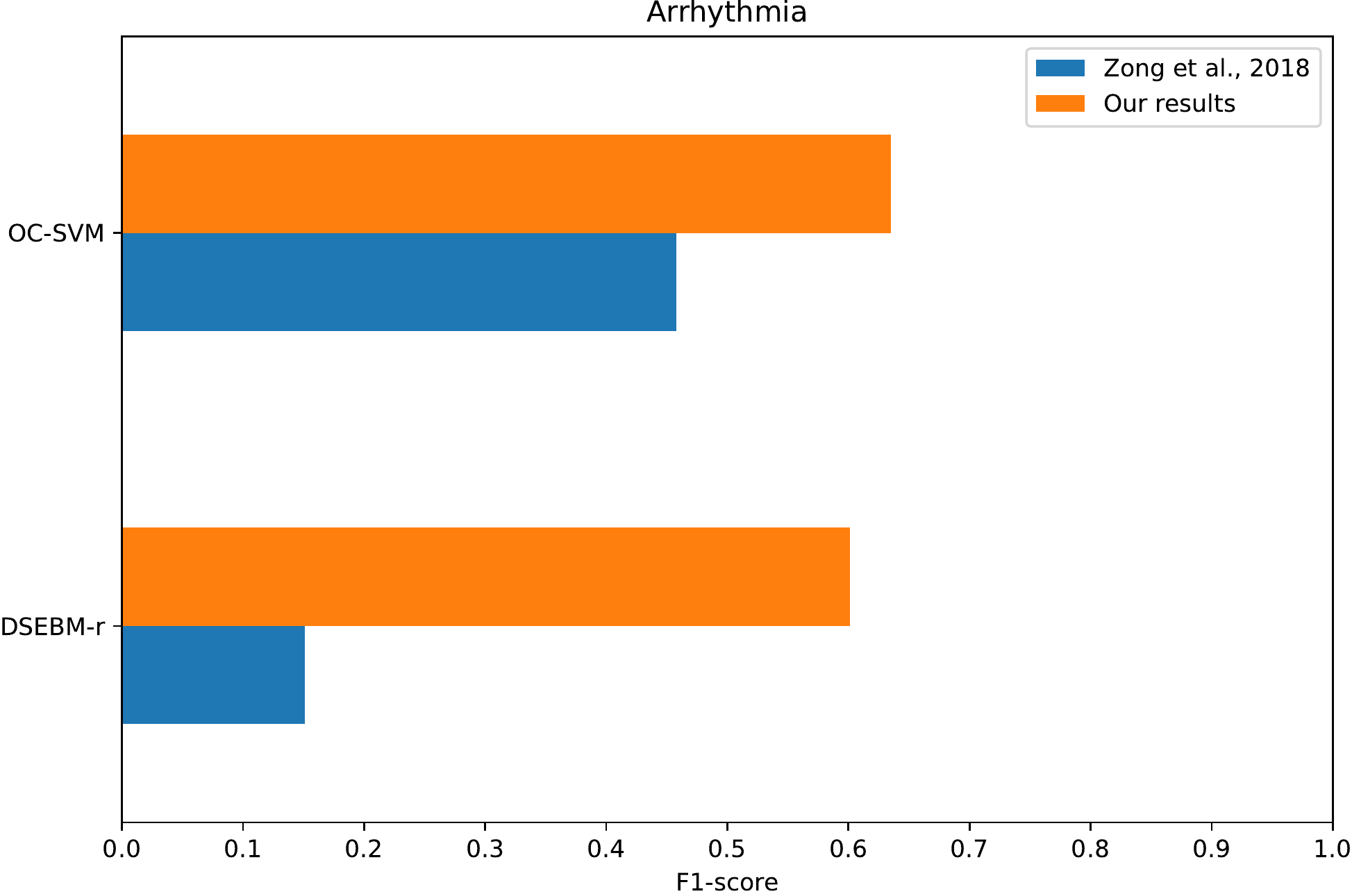}
         \caption{On Arrhythmia.}
        \label{dsebm_medical_difference_arrhythmia}
     \end{subfigure}
     \hfill
     \begin{subfigure}[b]{0.45\linewidth}
         \centering
         \includegraphics[width=\linewidth]{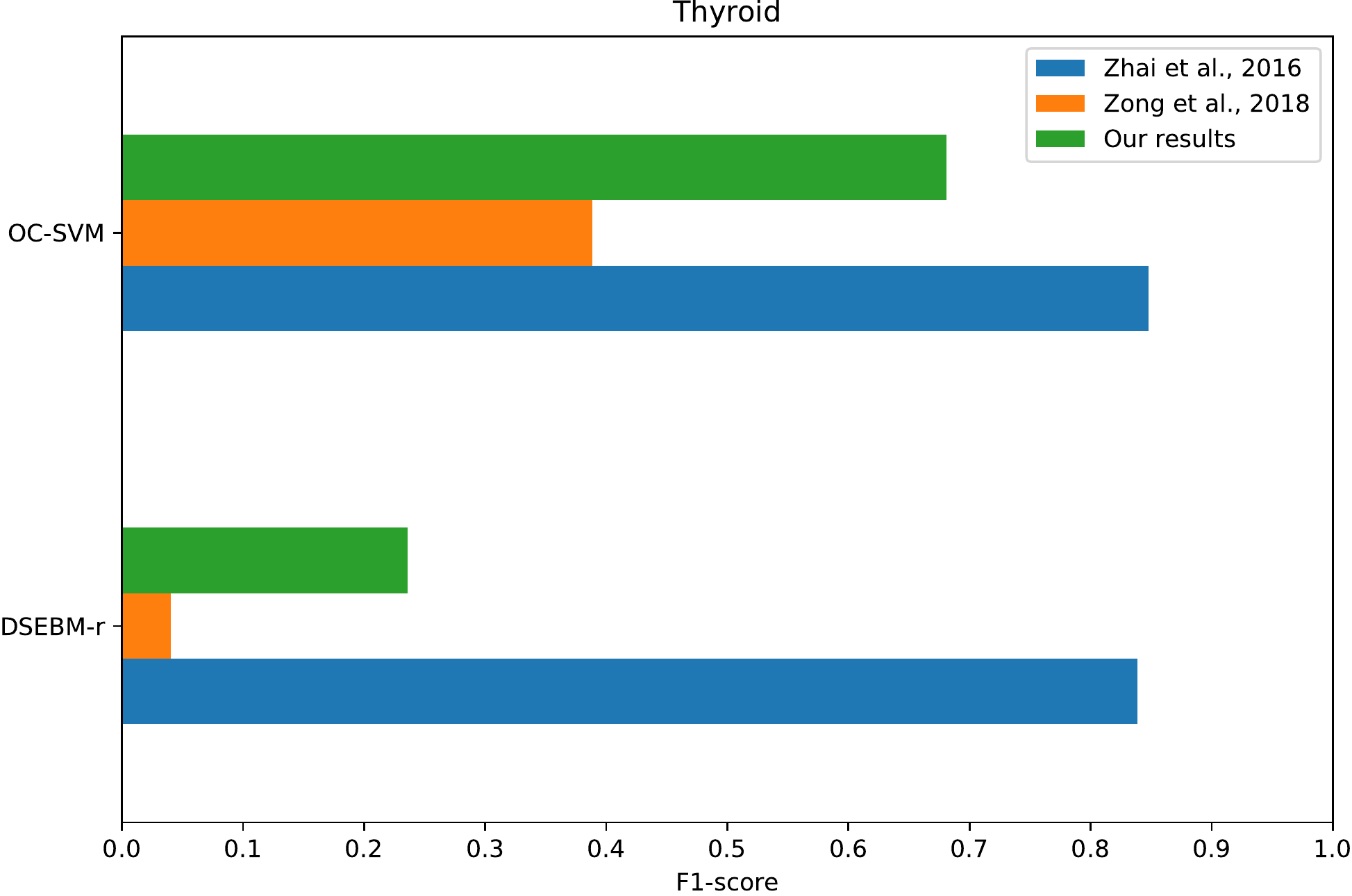}
        \subcaption{On Thyroid.}
        \label{dsebm_medical_difference_thyroid}
     \end{subfigure}
    \caption{Reported F1-scores for OC-SVM and DSEBM-r by different authors.}
\end{figure}

\textbf{Important differences with reported results.} We first note significant differences in reported precision, recall, and F1-scores between a few models. \cite{zong2018deep} report an F1-score of 0.0403 and 0.1510 for DSEBM-r on Arrhythmia and Thyroid respectively, while the original authors of the method \cite{zhai2016deep} obtained 0.8386 on Thyroid. Our results are quite different from both papers: 0.601 and 0.236 on Arrhythmia and Thyroid respectively. Results on baseline models also vary greatly. Figures \ref{dsebm_medical_difference_arrhythmia} and \ref{dsebm_medical_difference_thyroid} summarize these discrepancies. It’s unclear whether the experiments integrated all anomalies during testing or if some of them were left out during the subsampling phase. This would explain the large differences. \cite{zenati2018adversarially} and \cite{gong2019memorizing} cite the results from \cite{zong2018deep} on these models. Interestingly, the DSEBM implementation described in \cite{zenati2018adversarially}, available on Github\footnote{\href{ https://github.com/houssamzenati/Adversarially-Learned-Anomaly-Detection}{houssamzenati/Adversarially-Learned-Anomaly-Detection (github.com)}}, generates results very different from those reported in \cite{zenati2018adversarially}. They are, in fact, more in line with our scores.

\begin{figure}[!htbp]
     \centering
     \begin{subfigure}[b]{0.45\linewidth}
         \centering
         \includegraphics[width=\linewidth]{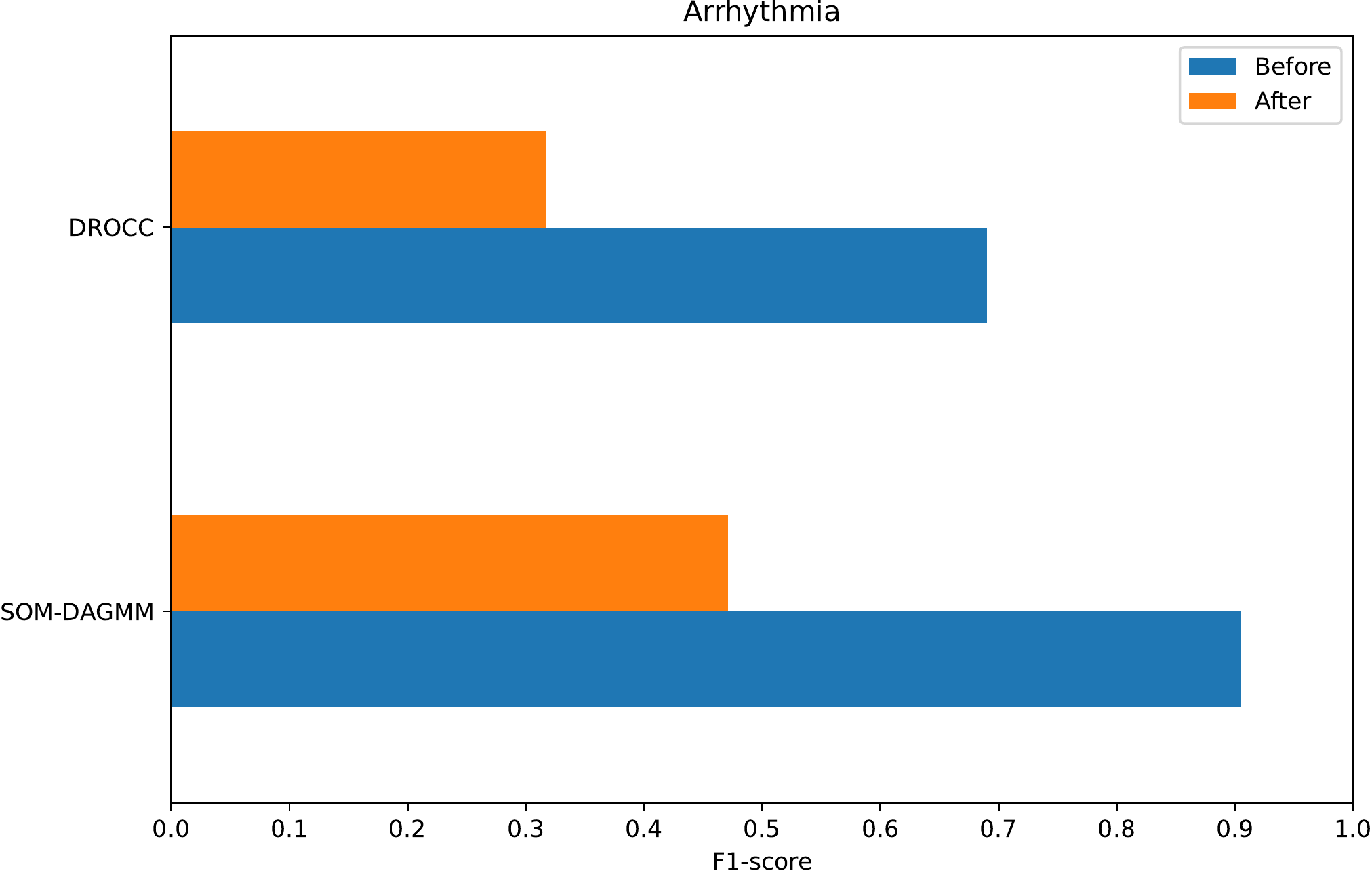}
        \caption{On the Arrhythmia dataset.}
        \label{drocc_medical_difference_arrhythmia}
     \end{subfigure}
     \hfill
     \begin{subfigure}[b]{0.45\linewidth}
         \centering
         \includegraphics[width=\linewidth]{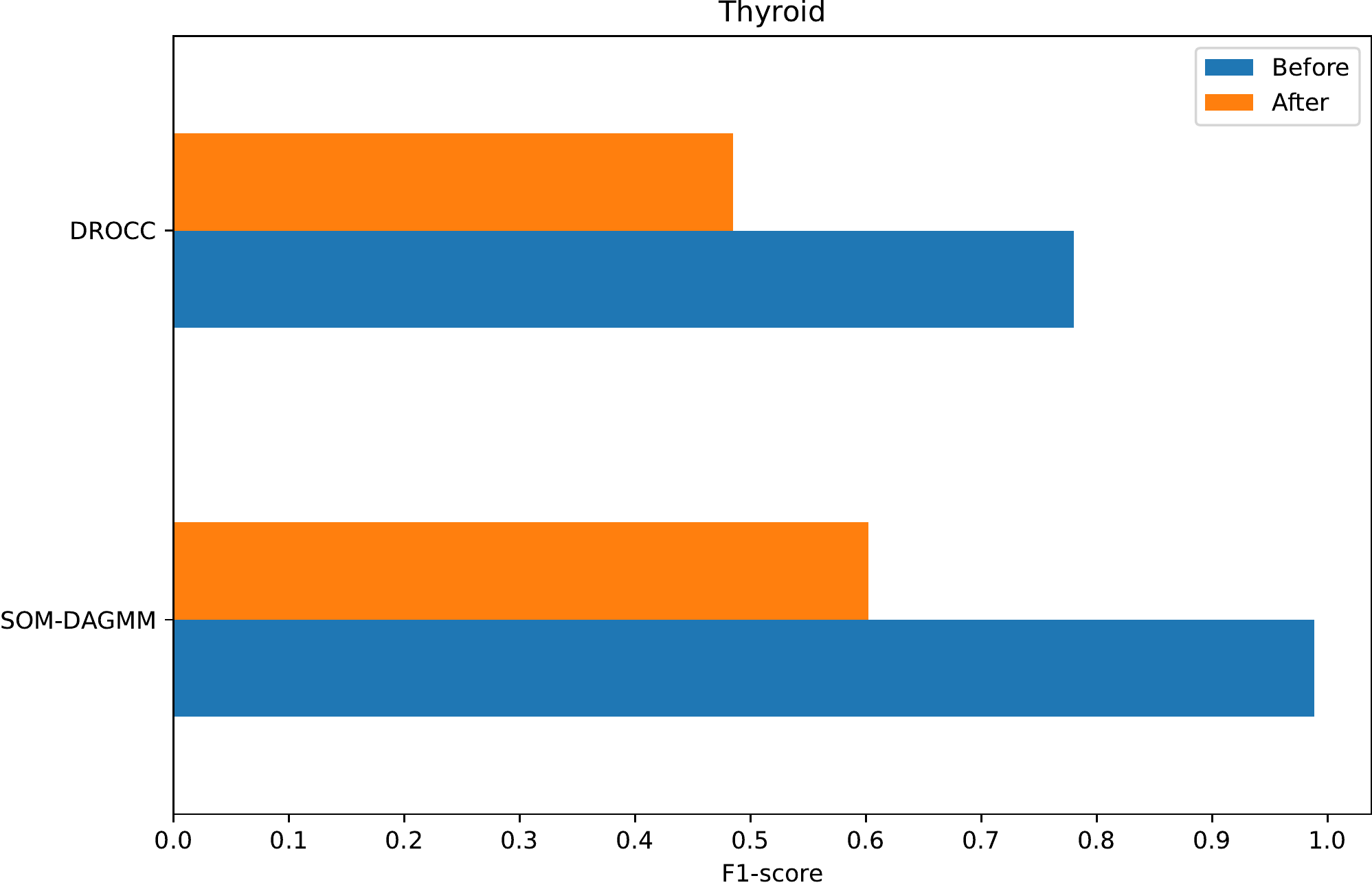}
        \caption{On the Thyroid dataset.}
        \label{drocc_medical_difference_thyroid}
     \end{subfigure}
    \caption{F1-score computed on different classes of interest.}
\end{figure}

\textbf{Taking the majority class as the class of interest yields overly optimistic results.} As displayed in Figures \ref{drocc_medical_difference_arrhythmia} and \ref{drocc_medical_difference_thyroid}, results on DROCC and SOM-DAGMM differ significantly when considering the minority class as the class of interest. Results drop from 0.78, 0.69 to 0.485 (-0.295) and 0.317 (-0.3729) on Thyroid and Arrhythmia respectively for DROCC. Similarly, our protocol generates 0.471 and 0.602 on the same datasets compared to 0.9053 (-0.4343) and 0.9888 (-0.3868) reported by SOM-DAGMM. Emphasizing predictions on the normal class can be misleading when dealing with skewness in class distribution. The probability of misclassification is much lower given their large number in the dataset. Conversely, anomalies are more challenging to detect as they are less frequent.

\textbf{Report both precision and recall scores for better interpretability.} Most research papers reviewed display both recall and precision scores, but too many still report only the F1-score. F1-score is the harmonic mean between precision and recall. Therefore, a good score can mask a poor recall with excellent precision or vice versa. For instance, DROCC’s excellent 0.996 recall score on CSE-CIC-IDS2018 indicates that the model doesn’t miss a lot of anomalies (low false-negative rate), but its 0.296 precision suggests it often flags normal samples as anomalies (high false-positive rate). This observation is not possible using only the F1-score (0.456).

\textbf{AUPR is more informative than AUROC.} The differences between AUROC and AUPR for all the models on the KDD and NSL-KDD are negligible. However, they differ significantly on CSE-CIC-IDS2018 where we see a significant drop (over 10\%) in performance between the two metrics. DAE drops from 0.882 to 0.689, DSEBM-r from 0.727 to 0.398 and ALAD from 0.856 to 0.615 to name a few. This further demonstrates our case that AUROC gives an optimistic perspective on a classifier’s performance by giving equal weights to predictions on abnormal and normal instances. In anomaly detection, we are interested in the performance on the anomalies and so it makes sense to use a metric more sensitive to predictions on that class. 

\textbf{Testing on KDD is insufficient.} KDD and its 10 percent variant NSL-KDD should only be used as a kind of basic sanity check since they do not provide distinctive insights into the performance of the methods. All implemented models -- both deep and shallow -- perform exceptionally well on this dataset, with F1-scores above 0.90. We can therefore conclude that KDD is highly trivial for anomaly detection. Instead, datasets such as CSE-CIC-IDS2018 should be preferred for comparison in the area of network intrusion detection, as most models report poor performance and results vary considerably on this more challenging dataset. Also, CSE-CIC-IDS2018 simulates a more realistic network, unlike KDD which was built over 20 years ago.

\textbf{Summary.} Among the surprising results, we note that our vanilla auto-encoder DAE outperforms more sophisticated reconstruction-based methods like DAGMM and MemAE on CSE-CIC-IDS2018. The large number of samples and intra-class variation on the same dataset could explain the downfall of DeepSVDD, DROCC, and one-class classification approaches in general. More generally, baseline methods with optimized hyper-parameters achieve more competitive F1-scores than reported in the literature so far. On Arrhythmia and Thyroid, they even outperform most of their deep-learning counterparts. NeuTraLAD, the transformation-based approach, offers consistently above-average performance across all datasets. The data-augmentation strategy is particularly efficient on small-scale datasets where samples are scarce. The only adversarial approach (ALAD) does not distinguish itself significantly from the other reconstruction-based methods, which is to be expected since it uses a reconstruction objective as its core.

\section{Conclusion}
In this paper, we emphasize the importance of using a reliable evaluation protocol to assess anomaly detection methods, revealed inconsistencies in reported results and claims, proposed a consistent evaluation protocol, and provided an updated evaluation of the twelve popular unsupervised anomaly detection methods on five widely used tabular datasets. The results reported in this paper give a better understanding of the current standing of the various unsupervised anomaly detection methods for tabular data.

We addressed the sources of inconsistencies in the evaluation protocols and offered a solution for each. We solve the data split problem by training on normal data only and using all of the anomalous samples in the test set. We discussed how the choices of performance metrics must be mindful of the imbalance in typical anomaly detection datasets. We advocate for the F1-score, precision, recall, and AUPR metrics to be reported to give a better picture of the performances of each method. The strategy for choosing the threshold for classification must be the same in all the evaluated methods, which we fixed to using the optimal threshold. We also mentioned how the choice of the class of interest can skew the evaluations and we set it to be the minority class. Finally, implementation details are vital to reproduce results.

We consider these results as preliminary and hope to extend our study to non-tabular data, especially time series and image datasets. We also hope that future work can compare their empirical results with ours following the same evaluation protocol and ultimately improve the consistency and reliability of future comparisons.

\bibliography{example_paper.bib}
\bibliographystyle{iclr2022_conference}

\appendix
\section*{Appendix}
\section{Models Description}
We chose a set of methods from highly cited papers published between 2016 and the time of this publishing and models considered as baseline. We chose baseline models based on the reputation of their paper, number of citations and number of times they are considered as baseline. We also include more recent models that claim state-of-the-art performance on a subset of the datasets previously described.

\textbf{Deep Structured Energy Based Models for Anomaly Detection \cite{zhai2016deep}}. DSEBM performs density estimation through energy-based models. The energy function is composed of neural networks and energy is accumulated across the multiple layers \cite{zenati2018adversarially}. Two different anomaly scores are studied: reconstruction error (DSEBM-r) and energy score (DSEBM-e).

\textbf{Deep Auto Encoding Gaussian Mixture Model \cite{zong2018deep}}. DAGMM trains an autoencoder and a feed-forward network in an end-to-end fashion. The reconstruction error and the latent representation produced by the autoencoder are given as input to a MLP which is used to estimate the parameters of a Gaussian Mixture Model (GMM). The output of the later network is ultimately used to compute the log-likelihood of the samples, which is then used as an anomaly score. While no official code is available for this model, the original paper provided enough information for us to reimplement it.

\textbf{Memory-augmented Deep Autoencoder \cite{gong2019memorizing}}. MemAE leverages the representational potential of an encoder with a memory module with a sparse attention-based addressing mechanism to record the prototypical patterns in the data. A decoder is used to reconstruct the original sample from the items retrieved from the memory module. High reconstruction errors are associated with anomalies. We reused the implementation offered by the authors on Github.

\textbf{Deep One-Class Classification \cite{ruff2018deep}}. DeepSVDD leverages the representational potential of deep neural networks to learn a representation of the data that encompasses a hypersphere. By minimizing the volume of this hypersphere,the network is encourage to extract the common factors of variation in the training data. Points outside the hypersphere are predicted as anomalies. Fortunately, a PyTorch implementation is made available by the authors.

\textbf{Deep Robust One-Class Classification \cite{goyal2020drocc}}. DROCC assumes that the normal points lie on a low-dimensional manifold that is well sampled. Based on this assumption, the authors develop a method that can train deep neural network architecture by generating anomalous points. For each normal point, gradient ascent is used to generate anomalous points that maximize the loss of the network. Using this adversarial approach, DROCC is able to synthetically generate data to train the DNN architecture in a supervised manner. We integrated the authors' PyTorch code (https://github.com/microsoft/EdgeML/).

\textbf{Aversarially Learned Anomaly Detection \cite{zenati2018adversarially}}. ALAD expands on the GAN foundations by adding an encoder to map data points to the latent space, and two more discriminators to ensure data-space and latent-space cycle-consistencies. ALAD uses the reconstruction error based on features extracted from an intermediate layer of a discriminator as the anomaly score. 

\textbf{Neural Transformation Learning for Deep Anomaly Detection Beyond Images \cite{qiu2021neural}}. NeuTraLAD leverages the recent success of contrastive learning in images and adapts it to tabular data. It uses a data augmentation scheme with a deterministic contrastive loss. This scheme encourages transformed samples to be similar to the original input while encouraging dissimilarity between the transformed samples. Instead of using predefined transformations (such as rotation, translation, cropping, etc.) that are well-suited for computer vision tasks, NeuTraLAD learns the data transformations with a set of neural networks. 

The previous SOTA methods are complemented with two of the most recurring baseline methods found within the anomaly detection literature.

\textbf{One-Class SVM \cite{scholkopf1999support}}. OC-SVM is a popular one-class classification SVM algorithm used for anomaly detection. We used the Scikit-Learn implementation and experimented on different $\nu$. The other parameters were set as defaults.

\textbf{Local Outlier Factor \cite{breunig2000lof}}. LOF classifies samples with substantially lower density than their neighbors as anomalies. The Scikit-Learn version was again used with optimized values for \texttt{n\_neighbors}.

\section{Hyperparameters}
In this appendix we describe the hyperparameters used to obtain the reported results.

\begin{table*}[!h]
    \tiny
    \centering
    \begin{tabular}{@{}lcccccccc@{}}
        \toprule & \multicolumn{5}{c}{ALAD} \\
        \cmidrule{2-6} & Batch & Epoch & Lat. dim. & Weight decay & Learning rate \\
        \hline
        Arrhythmia         & 128 & 10000 & 32      & 0.0001      & 0.0001\\
        Thyroid           & 128 & 20000 & 32       & 0.0001      & 0.0001\\
        KDDCUP 10         & 1024 & 100 & 32        & 0.0001      & 0.0001\\
        NSL-KDD           & 1024 & 200 & 32        & 0.0001      & 0.0001\\
        CSE-CIC-IDS2018   & 1024 & 150 & 32        & 0.0001      & 0.0001\\
        \bottomrule
    \end{tabular}
\caption{ALAD hyperparameters.}
\end{table*}

\begin{table*}[!h]
     \tiny
    \centering
    \begin{tabular}{@{}lccccc@{}}
        \toprule & \multicolumn{4}{c}{DAE} \\
        \cmidrule{2-5} & Batch & Epoch & Lat. dim. & Learning rate \\
        \hline
        Arrhythmia         & 128 & 10000 & 3   & 0.0001\\
        Thyroid           & 128 & 5000 & 2    & 0.0001\\
        KDDCUP 10         & 1024 & 100 & 2  & 0.0001\\
        NSL-KDD           & 1024 & 100 & 2   & 0.0001\\
        CSE-CIC-IDS2018   & 1024 & 100 & 2   & 0.0001\\
        \bottomrule
    \end{tabular}
\caption{DAE hyperparameters.}
\end{table*}

\begin{table*}[!h]
     \tiny
    \centering
    \begin{tabular}{@{}lccccc@{}}
        \toprule & \multicolumn{5}{c}{DAGMM} \\
        \cmidrule{2-6} & Batch & Epoch & Lat. dim. & Learning rate & Weight decay\\
        \hline
        Arrhythmia         & 128 & 10000 & 2   & 0.0001 & 0.0001\\
        Thyroid           & 128 & 5000 & 2    & 0.0001 & 0.0001\\
        KDDCUP 10         & 1024 & 200 & 1  & 0.0001 & 0.0001\\
        NSL-KDD           & 1024 & 200 & 1   & 0.0001 & 0.0001\\
        CSE-CIC-IDS2018   & 1024 & 100 & 1   & 0.0001 & 0.0001\\
        \bottomrule
    \end{tabular}
    \caption{DAGMM hyperparameters.}
\end{table*}

\begin{table*}[!h]
     \tiny
    \centering
    \begin{tabular}{@{}lccccc@{}}
        \toprule & \multicolumn{5}{c}{DSEBM} \\
        \cmidrule{2-6} & Batch & Epoch & Lat. dim. & Learning rate & Weight decay\\
        \hline
        Arrhythmia         & 128 & 10000 & 2   & 0.0001 & 0.0001\\
        Thyroid           & 128 & 5000 & 2    & 0.0001 & 0.0001\\
        KDDCUP 10         & 1024 & 100 & 512  & 0.0001 & 0.0001\\
        NSL-KDD           & 1024 & 100 & 512   & 0.0001 & 0.0001\\
        CSE-CIC-IDS2018   & 1024 & 100 & 512   & 0.0001 & 0.0001\\
        \bottomrule
    \end{tabular}
    \caption{DSEBM hyperparameters.}
\end{table*}

\begin{table*}[!h]
     \tiny
    \centering
    \begin{tabular}{@{}lcc@{}}
        \toprule & \multicolumn{2}{c}{DeepSVDD} \\
        \cmidrule{2-3} & Batch size & Number of output features\\
        \hline
        Arrhythmia         & 128   & 64\\
        Thyroid            & 128   &  1\\
        KDDCUP 10          & 1024   & 29\\
        NSL-KDD            & 1024   & 31\\
        CSE-CIC-IDS2018    & 1024   & 16\\
        \bottomrule
    \end{tabular}
\caption{DeepSVDD hyperparameters.}
\end{table*}

\begin{table*}[!h]
     \tiny
    \centering
    \begin{tabular}{@{}lccccccccc@{}}
        \toprule & \multicolumn{8}{c}{DROCC} \\
        \cmidrule{2-9} & Batch size & Threshold & Radius & $\mu$ & $\nu$ & Learning rate & Only CE epochs & Gradient ascent steps\\
        \hline
        Arrhythmia         & 256    & 70    & 16    & 0.5   & 0.1    & 0.01     & 50 &  50\\
        Thyroid            & 256    & 95    & 0.5   & 1.0   & 0.01   & 0.0001   & 50 &  50\\
        KDDCUP 10          & 1024   & 61    & 16    & 0.5   & 0.1    & 0.01     & 50 &  50\\
        NSL-KDD            & -1     & 35    & 16    & 0.5   & 0.1    & 0.01     & 50 &  50\\
        CSE-CIC-IDS2018    & 100    & 2     & 8.124 & 1.0   & 0.01   & 0.0001 & 50 &  50\\
        \bottomrule
    \end{tabular}
\caption{DROCC hyperparameters.}
\end{table*}

\begin{table*}[!h]
    \tiny
    \centering
    \begin{tabular}{@{}lccccccccccccc@{}}
        \toprule & \multicolumn{8}{c}{DUAD} \\
        \cmidrule{2-9} & Batch & Epoch & Lat. dim. & $r$ & $p_0$ & $p_s$ & Clusters & Learning rate \\
        \hline
        Arrhythmia         & 128 & 5000 & 3    & 10 & 35 & 30 & 8  & 0.0001 \\
        Thyroid           & 128 & 5000 & 2    & 10 & 35 & 30 & 8    & 0.0001  \\
        KDDCUP 10         & 1024 & 100 & 2    & 10 & 35 & 30 & 10    & 0.0001  \\
        NSL-KDD           & 1024 & 100 & 2    & 10 & 35 & 30 & 10    & 0.0001  \\
        CSE-CIC-IDS2018   & 1024 & 100 & 2    & 10 & 35 & 30 & 15   & 0.0001 \\
        \bottomrule
    \end{tabular}
\caption{DUAD hyperparameters.}
\end{table*}

\begin{table*}[!h]
    \tiny
    \centering
    \begin{tabular}{@{}lcccccccc@{}}
        \toprule & \multicolumn{6}{c}{MemAE} \\
        \cmidrule{2-7} & Batch & Epoch & Lat. dim. & Mem. dim. & Weight decay & Learning rate \\
        \hline
        Arrhythmia         & 128 & 10000 & 3    & 50    & 0.0001      & 0.0001\\
        Thyroid           & 128 & 20000 & 3    & 50    & 0.0001      & 0.0001\\
        KDDCUP 10         & 1024 & 200 & 3    & 50    & 0.0001      & 0.0001\\
        NSL-KDD           & 1024 & 200 & 3    & 50    & 0.0001      & 0.0001\\
        CSE-CIC-IDS2018   & 1024 & 50 & 3    & 250    & 0.0001      & 0.0001\\
        \bottomrule
    \end{tabular}
\caption{MemAE hyperparameters.}
\end{table*}

\begin{table*}[!h]
     \tiny
    \centering
    \begin{tabular}{@{}lcccccl@{}}
        \toprule & \multicolumn{5}{c}{NeuTraLAD} \\
        \cmidrule{2-7} & Batch & Epoch & Lat. dim. & Learning rate & Weight decay & Transformation type\\
        \hline
        Arrhythmia         & 128 & 200 & 32   & 0.0001 & 0.00001 & residual\\
        Thyroid           & 128 & 580 & 24    & 0.0001 & 0.00001 & residual\\
        KDDCUP 10         & 1024 & 40 & 32  & 0.0001 & 0.00001 & multiplicative\\
        NSL-KDD           & 1024 & 40 & 32   & 0.0001 & 0.00001 & multiplicative\\
        CSE-CIC-IDS2018   & 1024 & 25 & 32   & 0.0001 & 0.00001 & multiplicative\\
        \bottomrule
    \end{tabular}
\caption{NeuTraLAD hyperparameters.}
\end{table*}

\begin{table*}[!h]
     \tiny
    \centering
    \begin{tabular}{@{}lccccc@{}}
        \toprule & \multicolumn{4}{c}{SOM-DAGMM} \\
        \cmidrule{2-5} & Batch & Epoch & Lat. dim. & Learning rate \\
        \hline
        Arrhythmia         & 128 & 10000 & 2   & 0.0001\\
        Thyroid           & 128 & 5000 & 1    & 0.0001\\
        KDDCUP 10         & 1024 & 100 & 2  & 0.0001\\
        NSL-KDD           & 1024 & 100 & 2   & 0.0001\\
        CSE-CIC-IDS2018   & 1024 & 100 & 2   & 0.0001\\
        \bottomrule
    \end{tabular}
\caption{SOM-DAGMM hyperparameters.}
\end{table*}

\begin{table*}[!h]
     \tiny
    \centering
    \begin{tabular}{@{}lccccccc@{}}
        \toprule & \multicolumn{2}{c}{OC-SVM} &  \phantom{abc} & \multicolumn{2}{c}{LOF} \\
        \cmidrule{2-3} \cmidrule{5-6}  & Threshold & $\nu$ && Threshold & Number of neighbors \\
        \hline
        Arrhythmia         & 73    & 0.40   && 75    & 50\\
        Thyroid            & 97    & 0.05   && 96    & 20\\
        KDDCUP 10          & 78    & 0.25   && 77    & 100\\
        NSL-KDD            & 46    & 0.40   && 44    & 20\\
        CSE-CIC-IDS2018    & 86    & 0.01   && 88    & 15\\
        \bottomrule
    \end{tabular}
\caption{OC-SVM and LOF hyperparameters.}
\end{table*}

\end{document}